\documentclass{bmvc2k}
\newcommand{\figref}[1]{Fig.~\ref{#1}}
\newcommand{\secref}[1]{Section~\ref{#1}}

\newcommand{\tabref}[1]{Table~\ref{#1}}

\usepackage{hyperref}
\usepackage[all=normal,floats=tight]{savetrees}

\newcommand\blfootnote[1]{%
  \begingroup
  \renewcommand\thefootnote{}\footnote{#1}%
  \addtocounter{footnote}{-1}%
  \endgroup
}

\title{Self-supervised learning of a facial attribute embedding from video}

\addauthor{Olivia Wiles*}{ow@robots.ox.ac.uk}{1}
\addauthor{A. Sophia Koepke*}{koepke@robots.ox.ac.uk}{1}
\addauthor{Andrew Zisserman}{az@robots.ox.ac.uk}{1}

\addinstitution{
 Visual Geometry Group\\
 University of Oxford\\
 Oxford, UK
}
\runninghead{Wiles, Koepke, Zisserman}{Self-sup.\ facial attribute from video}

\def\etal{\emph{et al}\bmvaOneDot}
\def\networkname{FAb-Net}
\def\voxcombiname{VoxCeleb+}

\begin{document}

\maketitle

\begin{abstract}
\noindent 
\blfootnote{* Equal contribution.}
We propose a self-supervised framework for learning facial
attributes by simply watching videos of a
human face speaking, laughing, and moving over time.

To perform this task, we introduce  a network, \textbf{F}acial \textbf{A}ttri\textbf{b}utes-\textbf{Net} (\networkname),
that is trained to embed 
multiple frames from the same video face-track into a common low-dimensional space.

\noindent With this approach, we make three contributions: first, we show that the network can 
leverage information from multiple source frames by
predicting confidence/attention masks for each frame; second, we demonstrate that using a curriculum learning regime
  improves the learned embedding; finally,  we demonstrate
that the network learns a meaningful face embedding that encodes
information about head pose, facial landmarks and facial expression -- i.e. facial attributes --
{\em without} having been supervised with any labelled data. 
We are comparable or superior to state-of-the-art self-supervised methods on these tasks and approach the performance of supervised methods.

\end{abstract}

\section{Introduction}\label{sec:intro}
Babies and children are highly perceptive to the facial expressions of the people they interact with~\cite{gerull2002mother,ekman1979facial}.
The ability to understand and respond to changes in people's emotional state is similarly important for computer vision systems and affective systems when interacting with a human user.
Thus being able to predict head pose and expression is of vital importance.

Recently, leveraging deep learning has led to state-of-the-art results on a variety of tasks such as emotion recognition and facial landmarks detection.
Despite these advances, supervised methods require large amounts of labelled data which may be expensive or difficult to obtain in realistic, unconstrained settings,  or necessitate assigning data to ill-defined categories.
For example, categorising emotions with three human annotators leads to only 46\% agreement~\cite{barsoum2016training}, and labelling pose in the wild is notoriously difficult.
Moreover, performing each task independently does not leverage the fact that detecting landmarks requires understanding pose and facial features, which in turn correspond to expression.

\begin{figure}[h!]
\centering
\includegraphics[width=\linewidth]{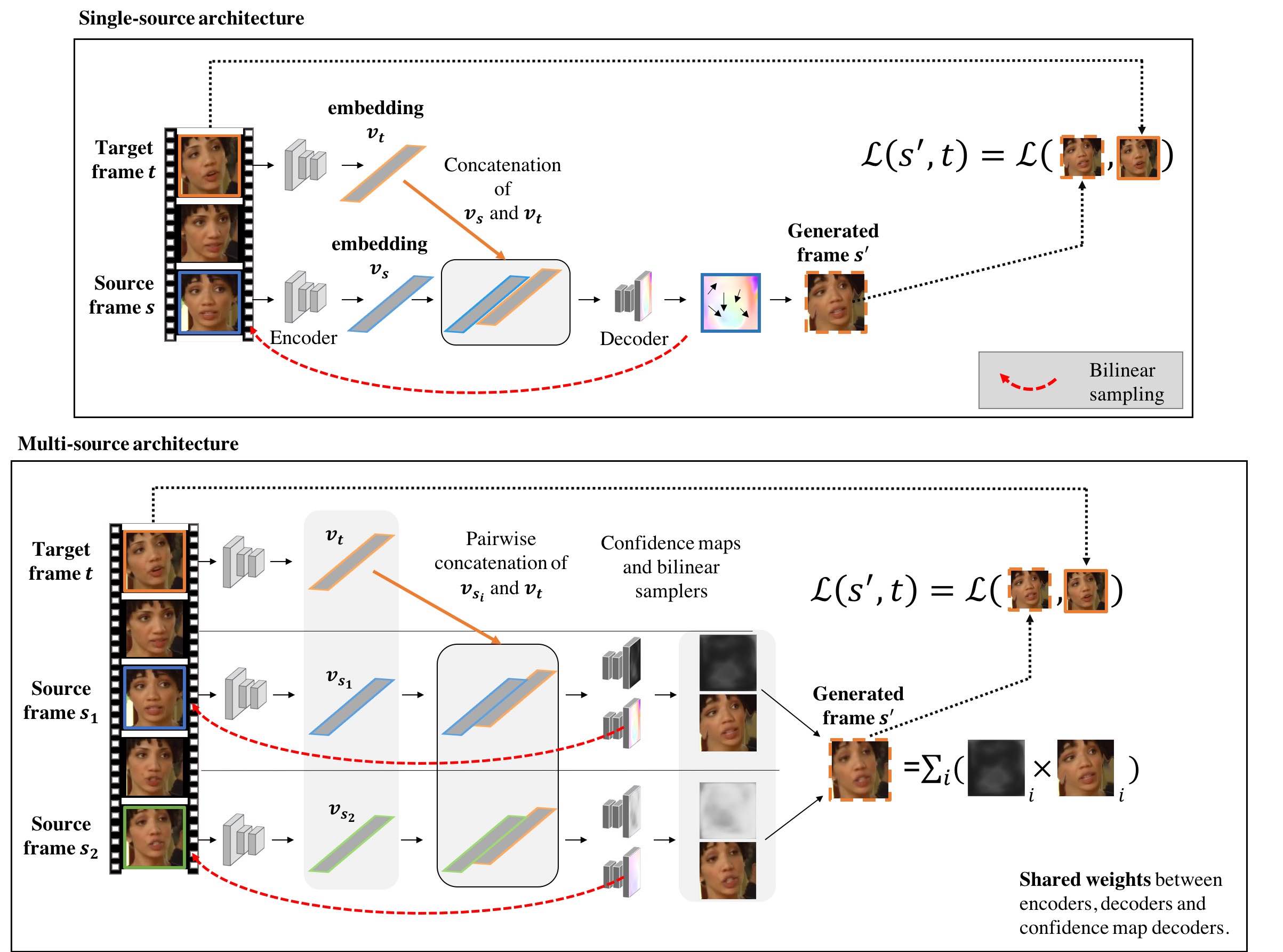}

\caption{An overview of \networkname. 
In the single-source case (top), the encoder-decoder architecture
takes one source frame and a target frame as inputs and learns to
generate the target frame. The $256$-dimensional outputs of the
encoders -- the source and target attribute embeddings -- are
concatenated and input to the decoder.  The decoder predicts
the point-to-point correspondence from the source frame to the target frame, and RGB values
for the generated frame are then obtained from the source frame using a bilinear sampler.
The network is
trained with an  $L1$ loss between the generated frame and the target
frame. In the multi-source case (bottom), the decoder also predicts a
confidence map, and the confidence maps are used to weight  the contributions of the different
source frames.}

\label{fig:overview}
\end{figure}

Consequently, we consider the following question: is it possible to learn an embedding of facial attributes that encodes landmarks, pose, emotion, etc.~in a self-supervised manner without {\em any} hand labelling?
The learned embedding can then be used for another task (e.g.~landmark, pose, and expression prediction) using a linear layer.
To do this, we contribute \networkname, a self-supervised framework for learning a low-dimensional face embedding for facial attributes (\secref{sec:method}).
We take advantage of video data which contains a large collection of images of the same person from different viewpoints and with varied expressions.
Given only the embeddings corresponding to a source and target frame, the
network is tasked  to map  the source frame to the target frame by predicting a flow field between them.
This proxy task forces the network to distil the information required to compute the flow field (e.g.\ the head pose and expression) into the source and target embeddings. 

After explaining the setup for a single source frame in
\secref{sec:singlesource}, we introduce our additional contributions:
a method for
leveraging multiple frames in order to improve the learned embedding
in \secref{sec:multisource};  and how a curriculum strategy for training
\networkname~in~\secref{sec:curriculum} can be used to improve performance.

The learned embedding is extracted and used for a variety of tasks such as landmark detection, pose regression, and expression classification in \secref{sec:experiments} by simply learning a linear layer.
Our results on these tasks are comparable or superior to other self-supervised methods, approaching the performance of supervised methods.
These experiments verify the hypothesis that our self-supervised framework learns to encode facial attributes which are useful for a variety of tasks.
Finally, the method is tested qualitatively by using the learned embedding to retrieve images with similar facial attributes across different identities.

\section{Related Work}\label{sec:related}

\noindent {\bf Self-supervised learning.}
Self-supervised methods such as \cite{Pathak16inp,Zhang16Col,doersch2015unsupervised,Noroozi16} require no manual labelling of the images;
instead, they directly use image data to provide proxy supervision for
learning good feature representations.  
The benefit of these approaches is that the  features learned using large quantities of available data can be transferred to other tasks/domains which have less or even no annotated data. 

To provide further supervision from image data, the images themselves can be transformed via a synthetic warp or rotation and the network trained to recognise the rotation~\cite{gidaris2018unsupervised} or to learn equivariant pixel embeddings \cite{Thewlis17a,Thewlis17b,novotny2018self}.

Of  more direct relevance to our training framework 
are self-supervised frameworks that use  video data~(\cite{Wang15,Fernando17ours,Misra16,lee2017unsupervised,jakab2018conditional,gan2018geometry,Wiles18a,xue2016visual,jia2016dynamic,
patraucean2015spatio,Chung17b,denton2017unsupervised,Agrawal15a,Zamir16}).

Our approach builds in particular on those that use frame
synthesis~\cite{xue2016visual,jia2016dynamic,patraucean2015spatio,denton2017unsupervised,Chung17b,Wiles18a},
though for us synthesis is a proxy task rather than the end goal.
Note, unlike~\cite{Wang15,Misra16,Fernando17ours,lee2017unsupervised},
we do not make use of the temporal ordering information inherent in a
video; nor do we predict future frames conditioned on a number of past
frames~\cite{patraucean2015spatio}, or explicitly predict the motion between frames as a
convolutional kernel~\cite{xue2016visual,jia2016dynamic}, or condition the generation on another modality
(e.g.\ voice~\cite{Chung17b}).  Instead, we treat the frames as an unordered set,  and
propose a simple formulation: that by embedding the source and target
frames in a common space and conditioning the transformation from
source to target frame on these embeddings, the learned embeddings
must learn to encode the relevant modes of variation necessary for the
transformation.

Concurrent to our work, Zhang~\etal~\cite{zhang2018unsupervised} and Jakab~\etal~\cite{jakab2018conditional} build on \cite{Thewlis17a} by
 using the discovered landmarks to reconstruct the original image.
However, unlike these works,  we do not place any constraints on the learned representation -- such 
as an explicit representation that encodes landmarks as heatmaps. 
\noindent {\bf Supervised learning of face embeddings.}
Given {\em known} (labelled) attribute information, e.g.\ for pose or expression, 
the embedding can be learned by training in a supervised
manner to directly predict the attribute~\cite{Kumar11,rudd2016moon,liu2015faceattributes}, 
 or to generate images (of faces, cars, or other classes) at a new,
{\em known} pose, expression,
etc.~\cite{Tran17drgan,yang2017neural,dosovitskiy2015learning,kulkarni2015deep,zhou2016view}.

 Another way of supervising a face embedding is to explicitly learn the parameters of a 3D morphable model (3DMM)~\cite{Blanz02}.
 As fitting a 3DMM is relatively expensive, \cite{Bas17,Tewari17} learn this end-to-end using either landmarks or a photometric error as supervision.
 However, unlike our method, these methods require either ground truth labels or a morphable model which fixes the modes of variation and the embedding.

\noindent An interesting half-way point is weak supervision, where the learned object or face embedding is conditioned for instance on object labels \cite{novotny2017anchornet} or weather/geo-location information~\cite{li2015two} respectively.
This requires additional meta-data, but results in embeddings that can represent attributes such as age and expression for faces or keypoints for objects.

\section{Method}\label{sec:method}

The aim is to train a network to learn an embedding that encodes facial attributes in a self-supervised manner, without any labels.
To do this, the network is trained to generate a target frame from one or multiple source frames by learning how to transform the source into the target frame. 
The source and target frames are taken from the same face-track of a person speaking, i.e.~the frames are  of the same identity but with different expressions/poses.
An overview of the architecture is given in \figref{fig:overview} and further described for a single source frame in \secref{sec:singlesource} and for multiple source frames in \secref{sec:multisource} (additional details are given in the supp.~material).

\subsection{Single-source frame architecture}
\label{sec:singlesource}
The input to the network is a source frame $s$ and a target frame $t$ from the same face-track.
These are passed through encoders with shared weights which learn a mapping $f$ from the input frames to a $256$-dimensional vector embedding (as shown in \figref{fig:overview}).
The embeddings corresponding to the target and source frames are
 $v_{t} = f(t)$ and $v_{s} = f(s)$ respectively.
The source and target embeddings are concatenated to give a $512$-dimensional vector which is upsampled via a decoder.
The decoder learns a mapping $g$ from the concatenated embeddings to a bilinear grid sampler, which samples from the source frame to create a new, generated frame $s' = g(v_t, v_{s}) (s)$.
Precisely, $g$ predicts offsets $(\delta x, \delta y)$ for each pixel location $(x,y)$ in the target frame; the generated frame $s'$ at location $(x,y)$ is obtained by sampling from the source frame $s$ according to these offsets : $s'{(x,y)} = s{(x+\delta x, y+\delta y)}$.
The network is trained to minimise the $L1$ loss between the generated and the target frame: $\mathcal{L}(s',t) = || t - s' ||_1$.

This setup enforces that the embeddings $v_s$ and $v_t$ represent facial attributes of the source and target frames respectively since the decoder maps  from the source frame $s$ to generate the frame $s'$ (i.e.\ it uses pixel RGB values from the source frame $s$ to create the generated $s'$ -- a similar formulation has been proposed concurrently to this work by \cite{vondrick2018tracking}). 

As the decoder is a function of the target and source attribute embeddings, and the decoder is the only place in the network where information is shared, the target attribute embedding must encode information about expression and pose
in order for the decoder to know where to sample
from in the source frame and where to place this information in the
generated frame.

\subsection{Multi-source frames architecture}
\label{sec:multisource}
While using two frames for training enforces that the network learns a high-quality embedding, additional source frames can be leveraged to improve the learned embedding.
This is achieved by also predicting a confidence heatmap -- a $1$ channel image --  for each source frame via an additional decoder.
The  heatmaps denote how confident the network is of the flow at each pixel location -- e.g.~if the source frame has a very different pose than the target frame, the confidence heatmap would have low certainty.
Moreover, it can express this for sub-parts of the image; if the mouth is closed in the source but open in the target frame, the confidence heatmap can express uncertainty in this region. 
The confidence heatmaps $C_i$ are combined pixel-wise for each source frame $s_i$ using a soft-max operation. 
For $n$ source frames, the loss function to be minimised is given as $\mathcal{L} = ||t - \frac{\sum_{i=1}^n e^{C_i} *  (g(v_t, v_{s_i})  (s_i)) }{\sum_{i=1}^n e^{C_i}} ||_1$.

\section{Curriculum Strategy}
\label{sec:curriculum}
The training of the network is divided into stages, so that knowledge can be built up over time as the examples given become progressively more difficult, as inspired by~\cite{bengio2009curriculum,Kumar10ours}. 
The loss computed by a forward pass is used to rank samples (i.e.~source and target frame pairs) in the batch according to their difficulty in a manner similar to~\cite{loshchilov2015online,simo2015discriminative,shrivastava2016training,nagrani2018learnable}.
However, these methods use only the most difficult samples, which was found to stop our network from learning. 
Similarly to~\cite{nagrani2018learnable}, using progressively more difficult samples proved crucial for the strategy's success.

Given a batch size of $N$ randomly chosen samples, i.e. source and target frame pairs, a forward pass is executed and the loss for each sample computed.
The samples are ranked and sorted according to this loss.
Initially the loss is back-propagated only on the samples in the batch which are in the 50th percentile (i.e.~the $0.5N$ samples with the lowest loss computed by the forward pass). These are assumed to be easier samples.
When the loss on the validation set plateaus, the subset to be back-propagated on is shifted by 10 (e.g.~the samples in the 10th-60th percentile range).
This is repeated 4 times until the samples being back-propagated on fall into the 40th-90th percentile range.
At this point the curriculum strategy is terminated, as it is assumed that the samples in the 90-100th range are too challenging or may be problematic (e.g.~there is a large shift in the background which is too challenging to learn).

\section{Experiments}\label{sec:experiments}

In this section, we evaluate the network and the learned embedding. In~\secref{exp:embeddingclassifier}, the performance of using \networkname's learned representation is compared to that of state-of-the-art self-supervised and supervised methods on a variety of tasks: facial landmark prediction, head pose regression and expression classification.

\secref{exp:additionalsourceframes} discusses the benefit  of using additional source frames,  and \secref{exp:informationretrieval} shows how the learned representation can be used for
retrieving images with similar facial attributes.

\noindent {\bf Training.} The model is trained on the VoxCeleb1 and VoxCeleb2 video datasets~\cite{Nagrani17,Chung18a}; we refer to the combined datasets as \voxcombiname. The \voxcombiname~dataset consists of videos of interviews containing more than 1 million utterances of around 7,000 speakers. The frames are extracted at 1 fps.
The frames are cropped, resized to $256\times256$, and the identities are randomly split into train/val/test (with a split of 75/15/10).

The models are trained in PyTorch~\cite{Pytorch} using SGD, an initial learning rate of $0.001$, and momentum $0.9$. 
When using the curriculum strategy described in \secref{sec:curriculum}, the batchsize is $N=32$, else $N=8$. 
The learning rate is divided by a factor of 10 when the loss on the validation set plateaus.
(If the curriculum strategy is used, the learning rate is updated only when the 40-90th percentile is considered.)
This is repeated until the loss converges. Further details about the training can be found in the supp. material.

\subsection{Using the embedding for regression and classification}
\label{exp:embeddingclassifier}
First, we investigate the representation learned in our embedding and evaluate whether it indeed encodes facial attributes by challenging it to predict three different attributes: landmarks, pose, and expression.

\noindent {\bf Setup.}
Given a network trained on \voxcombiname, a linear regressor or classifier is trained from the learned embedding to the output task.
The linear regressor/classifier consists of two layers: batch-norm~\cite{ioffe2015batch} followed by a linear fully connected layer with no bias.
The regression tasks are trained using a MSE loss.
The classification tasks are trained with a cross-entropy loss.
The parameters of the encoder are fixed while the two additional layers are trained on the training set of the target dataset using Adam~\cite{Kingma15}, a learning rate of $0.001$, $\beta_1=0.9$ and $\beta_2=0.999$.

\subsubsection{Baselines}
\noindent {\bf Self-supervised.}
There are prior publications on using self-supervision for landmark prediction on the datasets
we evaluate on, but none for predicting emotion on standard datasets.
Consequently, we implement an autoencoder and a set of state-of-the-art self-supervised methods~\cite{gidaris2018unsupervised,zhang2017split} for object detection and segmentation.
The baselines are trained using the same architecture as \networkname~but with their associated loss functions and training objectives.
For~\cite{zhang2017split}, the regression loss for both the L and ab channels is used.
These models are trained on \voxcombiname~until convergence, with the same training parameters and data augmentation as \networkname. More details are given in the supp.\ material.

\noindent {\bf VGG-Face descriptor.} 
We additionally compare to the {\em VGG-Face descriptor} which is obtained
from the 4096-dimensional FC7 features from a VGG-16 network trained on
the VGG-Face dataset \cite{Parkhi15}. 
Contrary to popular belief, it
has been recently shown that a network trained for 
identity does retain information about other facial attributes~\cite{colesynthesizing,ephrat2018looking}. We
use the VGG-Face descriptor to learn a linear
regression/classification layer to the desired attribute task.  This
provides a strong baseline, and the results obtained confirm the
finding that a network trained for identity does indeed encode
expression and to some extent also pose information.  However, note that unlike our
method, this face descriptor requires a large dataset of {\em
labelled} face images for training.

\subsubsection{Results}

\noindent {\bf Facial landmarks.}
\begin{table}
\begin{minipage}{0.51\linewidth}

\scriptsize
\centering 
\begin{tabular}{ | p{3.5cm} | c | c |}
\hline
{\bf Method}  & {\bf 300-W}  & {\bf MAFL} \\ \hline 
{\textit{\textbf{Self-supervised}}} & & \\
{\textit{Trained on VoxCeleb+}} & & \\
\networkname~& {6.31} & {3.78}\\ 
\networkname~w/ curric.~& {5.73} & 3.49 \\
\networkname~w/ curric., 3 source frames~& {\bf 5.71} & 3.44 \\ \hline
{\textit{Trained on CelebA}} & & \\
Jakab {\it et al.} \cite{jakab2018conditional} (2018) & -- & {\bf 3.08} \\
Zhang {\it et al.} \cite{zhang2018unsupervised} (2018)  & -- & {3.15} \\
Thewlis {\it et al.} \cite{Thewlis17a} (2017) & 9.30 & 6.67\\ 
Thewlis {\it et al.} \cite{Thewlis17b} (2017) & 7.97 & 5.83 \\ \hline \hline

{\textit{\textbf{Supervised}}} & & \\
{\textit{Trained on CelebA}} & & \\
MTCNN~\cite{zhang2014facial} (2014) & -- & {\bf 5.39} \\ 
LBF~\cite{ren2014face} (2014) &  6.32 & -- \\
CFSS~\cite{zhu2015face} (2015) & 5.76 & -- \\
cGPRT~\cite{lee2015face} (2015) & 5.71 & -- \\
DDN~\cite{yu2016deep} (2016) &  5.65 & -- \\
TCDCN~\cite{Zhang16} (2016) &  5.54 & -- \\ 
RAR~\cite{XiaoRobust} (2016) &  {\bf 4.94} & -- \\ \hline
VGG-Face descriptor~\cite{Parkhi15} & 11.16 & 5.92 \\ \hline

\end{tabular}

\caption{Landmark prediction error on 300-W and MAFL datasets. Lower is better.}	
\label{tab:300wmaflresults}
\end{minipage} \hspace{0.05cm}
\begin{minipage}{0.43\linewidth}
\centering
\scriptsize
\begin{tabular}{ | p{1.6cm} | c | c | c| c|}
\hline

{\bf Method} &  {\bf Roll}&  {\bf Pitch} & {\bf Yaw} & {\bf MAE}\\ \hline 
{\textit{\textbf{Self-supervised}}} & & & & \\
\networkname~ & {{5.54}}$^{\circ}$ & {7.84}$^{\circ}$ & {12.93}$^{\circ}$ & {8.77}$^{\circ}$\\ 
\networkname~w/ curric. & {{5.33}}$^{\circ}$ & {7.21}$^{\circ}$ & {11.34}$^{\circ}$ & {7.96}$^{\circ}$\\ 
\networkname~w/ curric., 3 source frames & {\bf{5.14}}$^{\circ}$ & {\bf 7.13}$^{\circ}$ & {\bf 10.70}$^{\circ}$ & {\bf 7.65}$^{\circ}$\\ \hline \hline
{\textit{\textbf{Supervised}}} & & & & \\
VGG-Face descriptor~\cite{Parkhi15} & {\bf 8.24}$^{\circ}$  & {8.36}$^{\circ}$ & {18.35}$^{\circ}$ & {11.65}$^{\circ}$ \\
KEPLER \cite{Kumar17} (2017) & {8.75}$^{\circ}$  & {\bf 5.85}$^{\circ}$ & {\bf 6.45}$^{\circ}$ & {\bf{7.02}}$^{\circ}$ \\ \hline
\end{tabular}
\caption{Pose prediction error on the AFLW test set from \cite{Kumar17}. Lower is better.}	
\label{tab:poseresults}
\end{minipage}

\end{table}
Facial landmark locations are regressed from the learned embedding and compared to state-of-the-art methods on MAFL~\cite{Zhang16} and the more challenging $300$-W~\cite{sagonas2016300} datasets.
The evaluation is performed as outlined in~\cite{Zhang16,Thewlis17a}, and the errors given in inter-ocular distance. 
For MAFL, 5 facial landmarks are regressed for 19k/1k train/test images. 
For $300$-W, $68$ landmarks are regressed for $3148/689$ train/test images which are obtained (as described in~\cite{Thewlis17a}) from combining multiple datasets~\cite{zhu12facewild,Belhumeur13,Zhou13workshops}.

\begin{figure}
\centering
\includegraphics[width=0.1\linewidth]{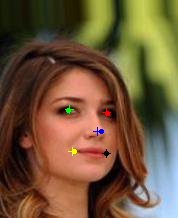}
\includegraphics[width=0.1\linewidth]{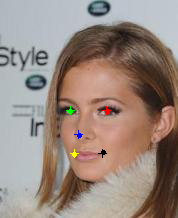}
\includegraphics[width=0.1\linewidth]{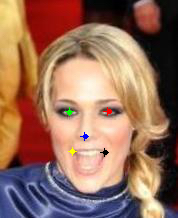}
\includegraphics[width=0.1\linewidth]{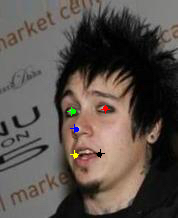}
\includegraphics[width=0.1\linewidth]{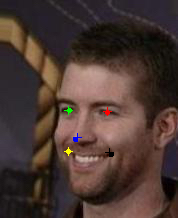}
\includegraphics[width=0.1\linewidth]{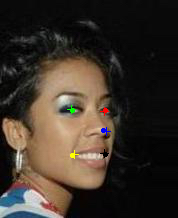}
\includegraphics[width=0.1\linewidth]{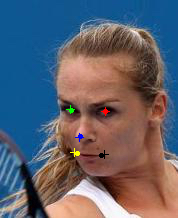}
\includegraphics[width=0.1\linewidth]{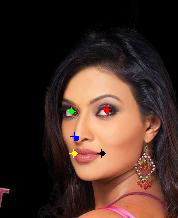}
\includegraphics[width=0.1\linewidth]{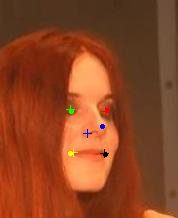}
\caption{Landmark prediction visualisation for \networkname~on the MAFL dataset. A dot denotes ground truth and the cross \networkname's prediction. A failure case is shown to the right.}
\label{fig:aflwresults}
\end{figure}

\begin{figure}
\centering
\subfigure[Ground truth landmarks]{
\centering
\includegraphics[height=0.1\linewidth]{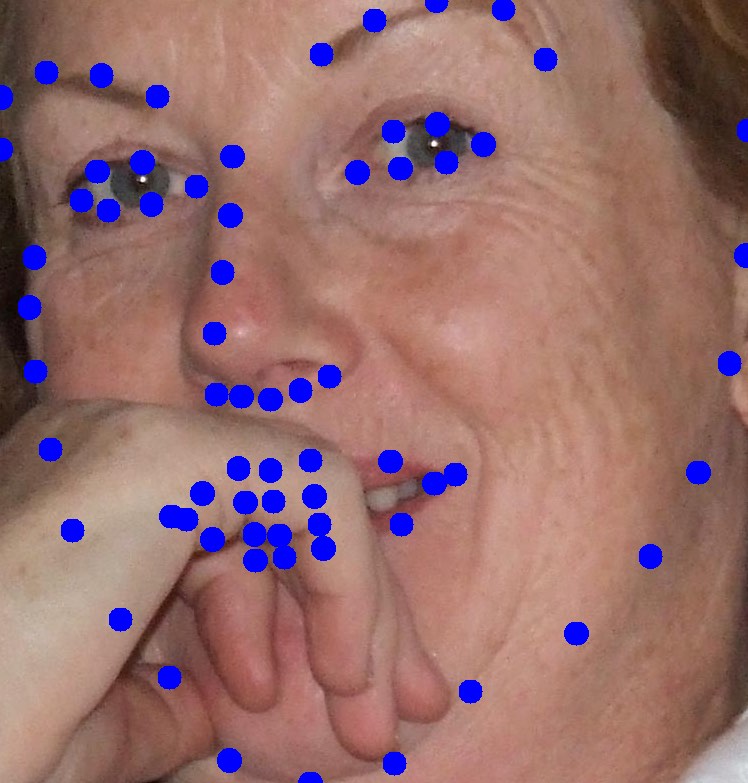}
\includegraphics[height=0.1\linewidth]{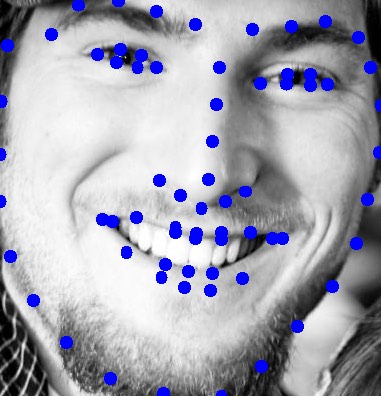}
\includegraphics[height=0.1\linewidth]{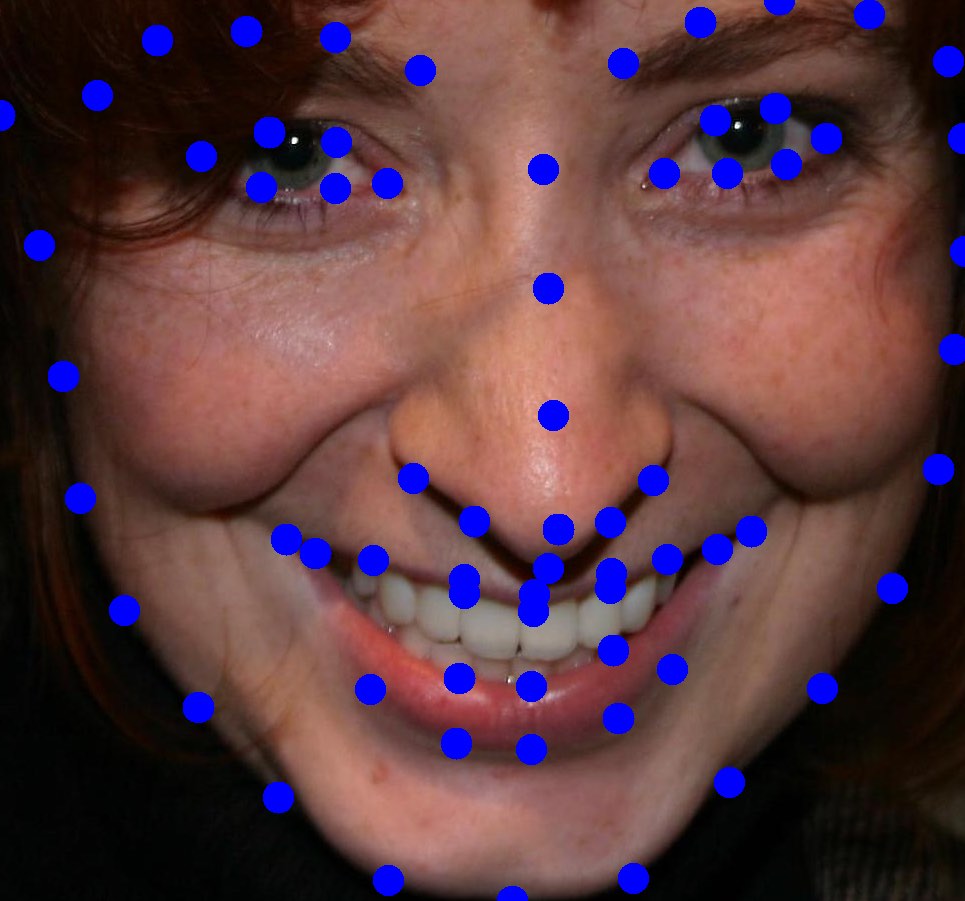}
\includegraphics[height=0.1\linewidth]{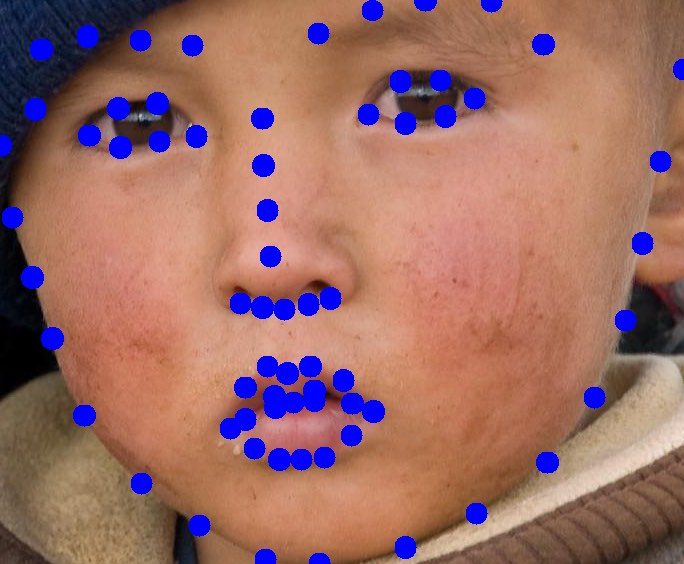}
\includegraphics[height=0.1\linewidth]{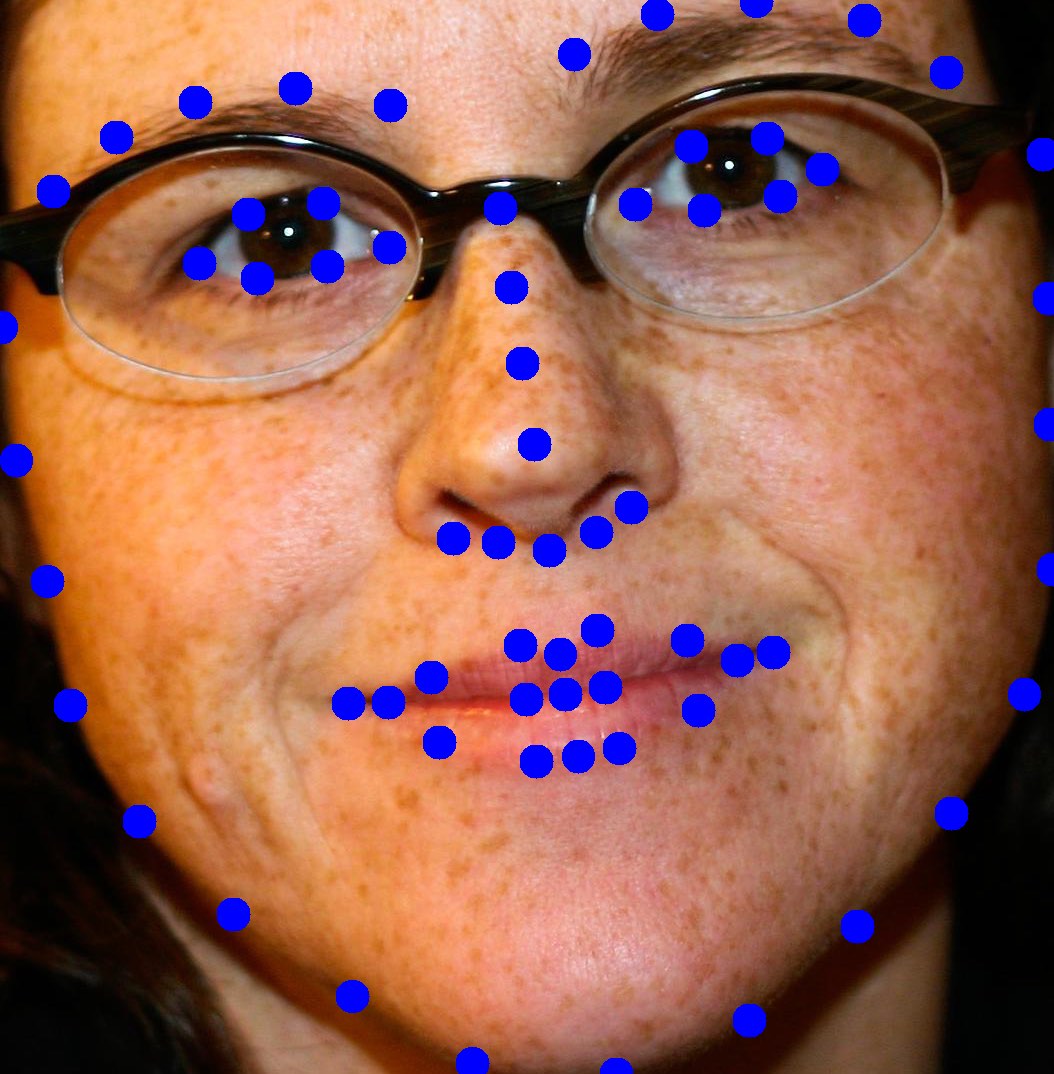}
\includegraphics[height=0.1\linewidth]{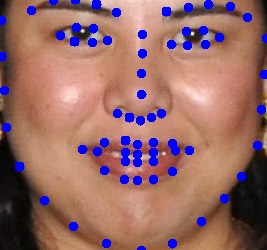}
\includegraphics[height=0.1\linewidth]{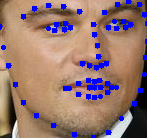}
\includegraphics[height=0.1\linewidth]{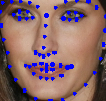}
\includegraphics[height=0.1\linewidth]{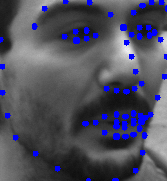}
}

\subfigure[\networkname's predicted landmarks]{
\centering
\includegraphics[height=0.1\linewidth]{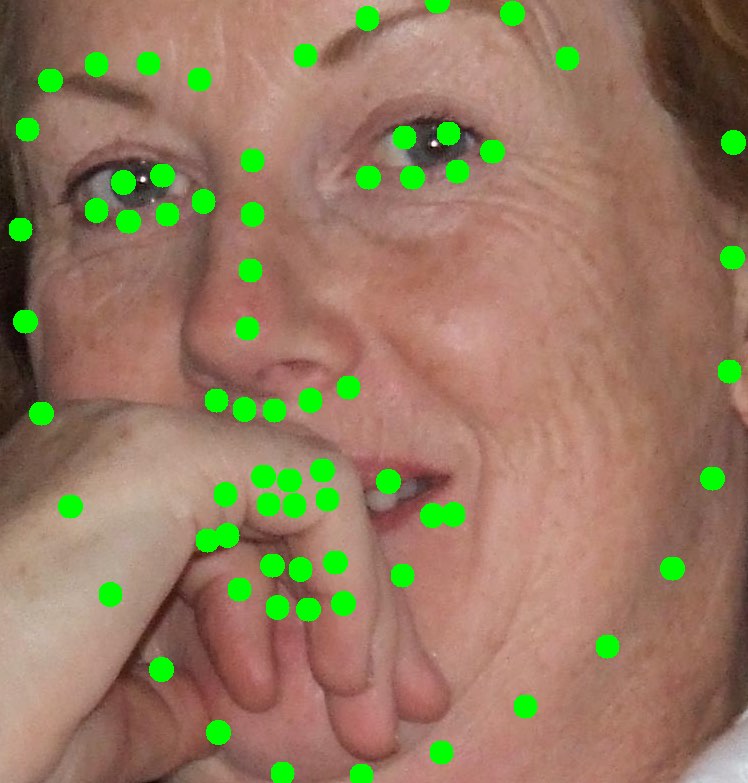}
\includegraphics[height=0.1\linewidth]{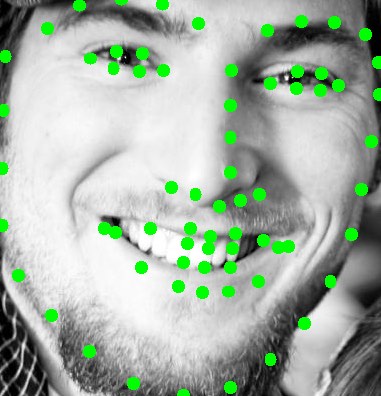}
\includegraphics[height=0.1\linewidth]{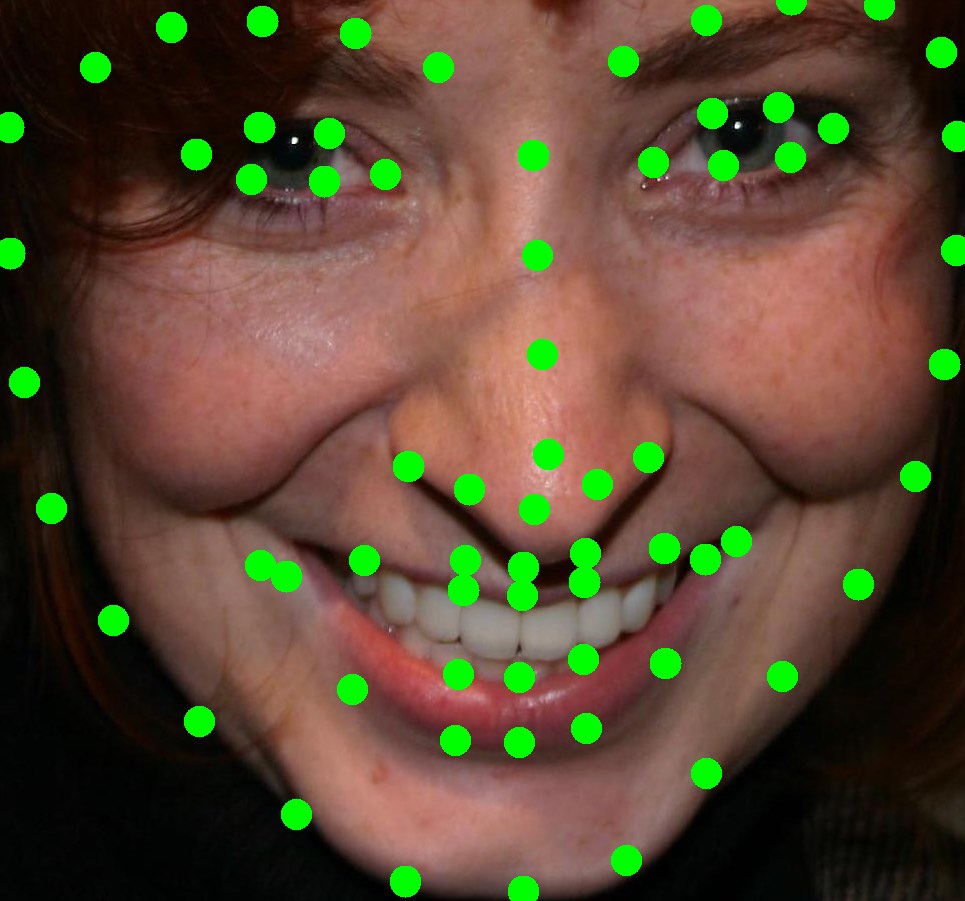}
\includegraphics[height=0.1\linewidth]{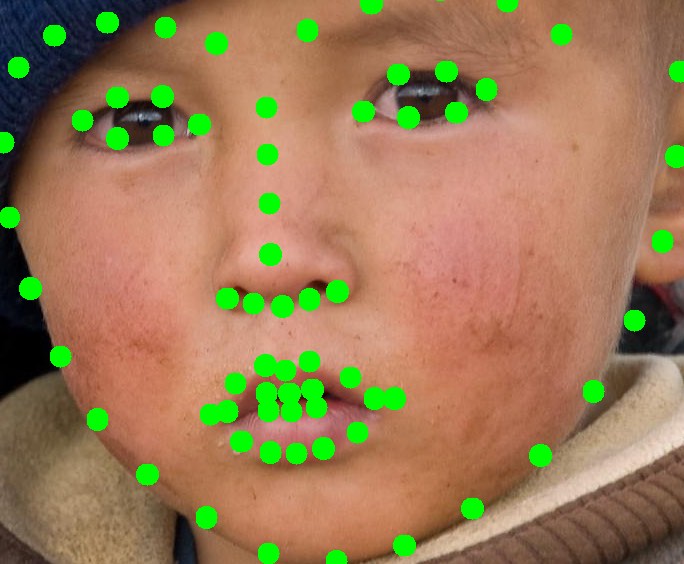}
\includegraphics[height=0.1\linewidth]{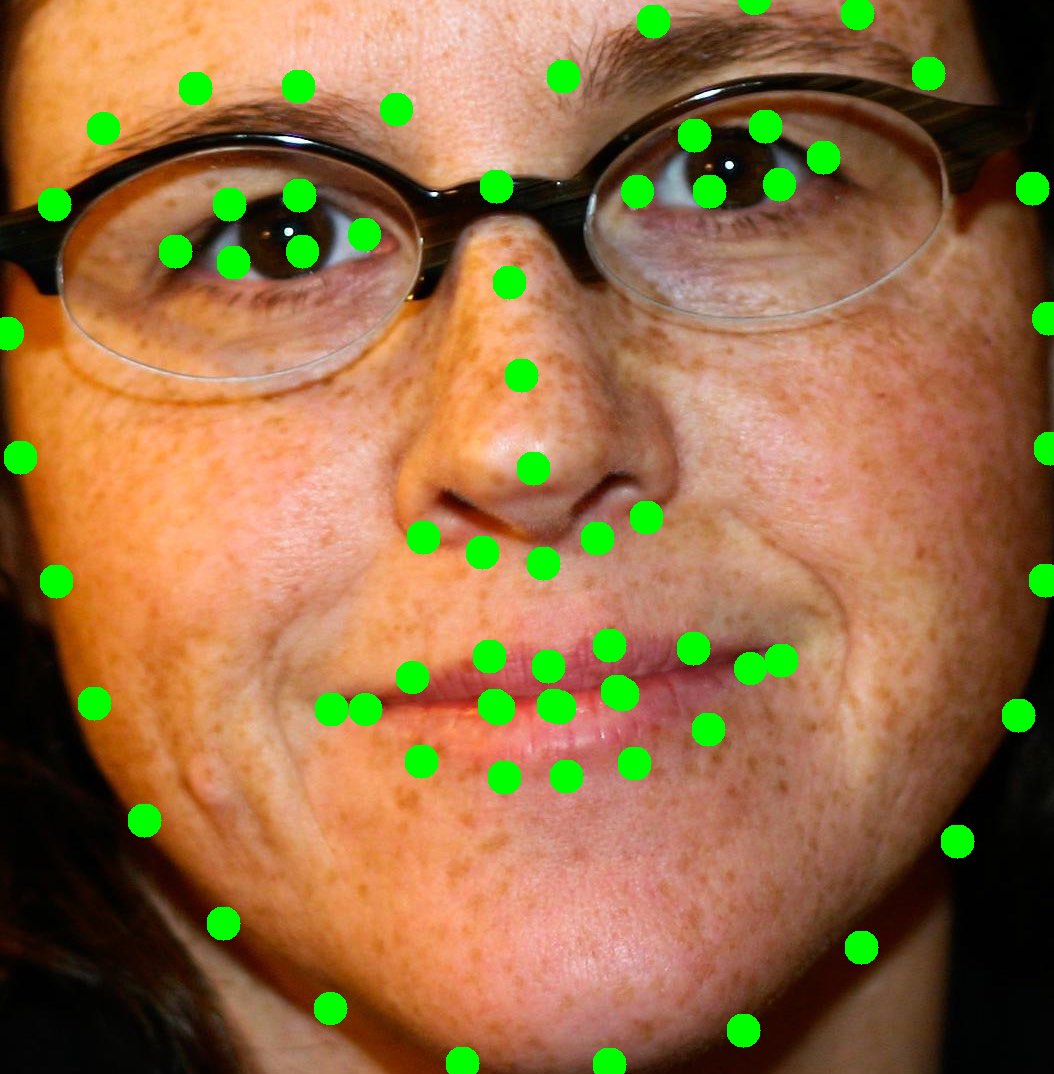}
\includegraphics[height=0.1\linewidth]{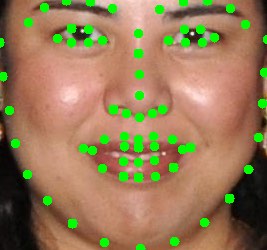}
\includegraphics[height=0.1\linewidth]{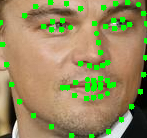}
\includegraphics[height=0.1\linewidth]{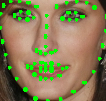}
\includegraphics[height=0.1\linewidth]{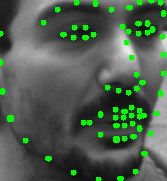}
}
\caption{Landmark prediction visualisation for \networkname~on the $300$-W dataset.}
\label{fig:300wresults}
\end{figure}

The results are reported in Table~\ref{tab:300wmaflresults} and some qualitative results are visualised in \figref{fig:aflwresults} and \figref{fig:300wresults}. 
These results first demonstrate that fine-tuning with additional views and our curriculum strategy improve the embedding learned by \networkname.
Second, these results show that our method performs competitively or better than state-of-the-art unsupervised landmark detection methods, better than the VGG-Face descriptor baseline and competitively with state-of-the-art supervised methods.
This is achieved even though the other self-supervised methods~\cite{zhang2018unsupervised,Thewlis17a,Thewlis17b,jakab2018conditional} are explicitly engineered to detect landmarks whereas our method is not. In addition to that, our method is able to bridge the domain gap between VoxCeleb+ and CelebA~\cite{liu2015faceattributes} -- the other self-supervised methods that we compare to are pre-trained on CelebA.

\noindent \noindent {\bf Pose.}
The learned embedding is used for pose prediction and compared to a supervised method \cite{Kumar17} and to using the VGG-Face descriptor. 
To perform the evaluation, the linear regression is trained from the given embedding to head pose labels using the AFLW dataset \cite{Koestinger11}, but after leaving out the $1,000$ images of the AFLW test set from \cite{Kumar17}.
 As can be seen in \tabref{tab:poseresults}, \networkname~performs better in predicting the roll angle, and the MAE is comparable to \cite{Kumar17} which is supervised with head pose labels. 
Furthermore, our embedding outperforms the VGG-Face descriptor which is trained on identities; i.e.~our learned embedding encodes more information about head pose.

\noindent {\bf Expression.}
\begin{table}
\scriptsize
\centering
\begin{tabular}{ | p{2cm} | c | c | c | c | c | c | c | c | c | c | c || c |}
\hline
 &  \multicolumn{11}{|c||}{AUC for different AUs} &\\ \hline
 & 1 & 2 & 4 & 5 & 6 & 9 & 12 & 17 & 20 & 25 & 26 & avg. \\ \hline 
 {\em \bf Self-supervised} & & & & & & & & & & & & \\
 \networkname~& {72.0} &  {68.9} &  {73.2} &  {69.4} &  {88.2} &  {78.6} &  {89.5} &  {71.0} &  {75.9} &  {81.4} &  {72.0} &  {76.4} \\ 
 \networkname~w/ curriculum& {73.4} &  {71.8} &  {75.3} &  {67.8} &  {90.4} &  {78.8} &  {91.9} &  {72.4} &  {74.5} &  {\bf 83.7} &  {73.3} &  {77.6} \\ 
 \networkname~w/ curriculum, 3 source frames & {\bf 74.1} &  {\bf 72.3} &  {\bf 75.8} &  {\bf 68.8} &  {\bf 90.7} &  {\bf 81.8} &  {\bf 92.5} &  {\bf 73.7} &  {\bf 77.2} &  {83.6} &  {\bf 73.6} &  {\bf 78.6} \\ \hline
 Gidaris \etal \cite{gidaris2018unsupervised} & 68.6 & 64.0 & 72.8 & 70.0 & 83.9 & 78.1 & 83.8 & 68.4 & 72.6 & 73.1 & 67.2 & 72.9 \\ 
 SplitBrain \cite{zhang2017split} & 65.5 & 59.8 & 66.7 & 60.8 & 71.8 & 65.8 & 73.3 & 64.5 & 57.4 & 68.1 & 61.1 & 65.0 \\
  Autoencoder & 67.2 & 60.5 & 70.1 & 65.1 & 79.6 & 70.4 & 80.1 & 68.3 & 66.5 & 70.5 & 64.1 & 69.3 \\ \hline \hline
  {\em \bf Supervised} & & & & & & & & & & & & \\
VGG-Face descriptor~\cite{Parkhi15} & {\bf 81.8} &  {\bf 83.0} &  {83.5} &  {81.8} &  {92.0} &  {\bf 90.9} &  {95.7} &  {\bf 80.6} &  {85.2} &  {86.5} &  {73.0} &  {84.9} \\ 
VGG-11 (from scratch) & 74.7 &  77.2 &  {\bf 85.8} &  {\bf 83.7} &  {\bf 93.8} &  89.7 &  {\bf 97.5} &  78.3 &  {\bf 86.9} &  {\bf 96.4} &  {\bf 81.5} &  {\bf 86.0} \\ \hline
\end{tabular}
\caption{Expression classification results for state-of-the-art self-supervised and supervised methods on EmotioNet~\cite{benitez2017emotionet} for multiple facial action units (AUs). Higher is better for AUC.}

\label{tab:emotionetresults}
\end{table}
\begin{table}
\scriptsize
\centering
\begin{tabular}{ | p{2.6cm} | c | c | c | c | c | c | c | c || c |}
\hline
 &  \multicolumn{9}{|c|}{AUC} \\ \hline
 & Neutral & Happy & Sad & Surprise & Fear & Disgust & Anger & Contempt & avg. \\ \hline 
{\textit{\textbf{Self-supervised}}} & & & & & & & & &  \\
 \networkname~ &  {70.0} &  {87.6} &  68.8 &  {75.5} &  {76.5} &  {70.0} &  {73.2} &  {71.2} &  {74.2} \\ 
\networkname~w/ curric. & 71.5 & 90.0 &  70.8 &  78.2 &  77.4 &  72.2 &  75.7 &  72.1 &  76.0  \\ 
  \networkname~w/ curric., 3 source frames &  {\bf 72.3} & {\bf 90.4} &  {\bf 70.9} &  {\bf 78.6} & {\bf 77.8}  &  {\bf 72.5} &  {\bf 76.4} &  {\bf 72.2} &  {\bf 76.4} \\ \hline
Gidaris \etal \cite{gidaris2018unsupervised} & 67.8 & 84.9 & {69.0} & 73.9 & 75.7 & 69.8 & 71.5 & 68.7 & 72.7 \\  
 SplitBrain \cite{zhang2017split} & 63.9 & 74.7 & 64.2 & 61.3 & 68.3 & 58.6 & 68.2 & 62.8 & 65.3 \\
Autoencoder & 65.8 & 80.0 & 64.7 & 66.1 & 70.6 & 63.4 & 68.3 & 65.0 & 68.0 \\ \hline \hline 
{\textit{\textbf{Supervised}}} & & & & & & & & &\\
VGG-Face descriptor~\cite{Parkhi15} & 75.9 & 92.2 & 80.5 & 81.4 & 82.3 & 81.4 & 81.2 & 77.1 & 81.5 \\
 AlexNet~\cite{mollahosseini2017affectnet} & -- & --    & --     & --	   & 	--    & --	& --	&  -- & {\bf 82} \\ \hline
\end{tabular}
\caption{Expression classification results for state-of-the-art self-supervised and supervised methods on AffectNet~\cite{mollahosseini2017affectnet}. Higher is better for AUC.}

\label{tab:affectnetresults}
\end{table}
We evaluate the performance of our learned embedding for expression estimation on two datasets: AffectNet~\cite{mollahosseini2017affectnet} and EmotioNet~\cite{benitez2017emotionet}, which both contain over 900,000 images.
These datasets are taken `in-the-wild' as opposed to in a constrained environment.
AffectNet contains 8 facial expressions (neutral, happy, sad, surprise, fear, disgust, anger, contempt) and EmotioNet contains 11 action units (AUs) (combinations of AUs correspond to facial expressions).

Both of these datasets were organised for challenges with a held out, unreleased test set.
Therefore, the train set is subdivided into two subsets; one is used for training and the other for validation.
The validation set of the original dataset is used to test the different models.
The linear classifier for EmotioNet is trained with a binary cross-entropy loss for each AU, whereas for AffectNet, a cross-entropy loss is used.
Both training datasets are highly imbalanced.
As a result, the examples from the under-represented classes are re-weighted inversely proportionally to the class frequencies to penalise the loss more heavily for mis-classifying images of the under-represented classes.

The embedding learned by \networkname~is compared to a number of self-supervised and supervised methods by measuring the Area Under the ROC curve (AUC).
For each class (e.g.~emotion or AU), the AUC is computed independently and the result is averaged over all classes.
The results are reported in \tabref{tab:emotionetresults} and \tabref{tab:affectnetresults} for EmotioNet and AffectNet respectively showing that our network performs better than other self-supervised methods over both metrics when given the same training data. 
This is supposedly due to the fact that the network must learn to transform the source frame in order to generate the target frame.  
As parts of the face move together (e.g.~an eyebrow raise or the lips when the mouth opens), the embedding must learn to encode information about facial features and thereby encode expression.
Interestingly, the autoencoder performs well, presumably due to the restricted nature of this domain.

\networkname~is also not far off supervised methods despite the domain shift; \voxcombiname~consists only of people being interviewed (and consequently with mostly neutral/smiling faces), so it does not include the range/extremity of expressions found in AffectNet or EmotioNet.
Finally, it can be observed that the VGG-Face descriptor trained to predict identities does surprisingly well at predicting emotion.

\noindent {\bf Discussion.} \networkname~has achieved impressive performance, as most self-supervised methods when transferred to another task have a large gap in comparison to supervised methods.
There is no gap or a small gap for the smaller datasets (landmarks/pose) and the model approaches supervised performance for the larger datasets (expression). 

\begin{figure*}[t]
    \centering

\begin{minipage}[t][0.56\width]{0.4\textwidth}%
    \includegraphics[width=\linewidth]{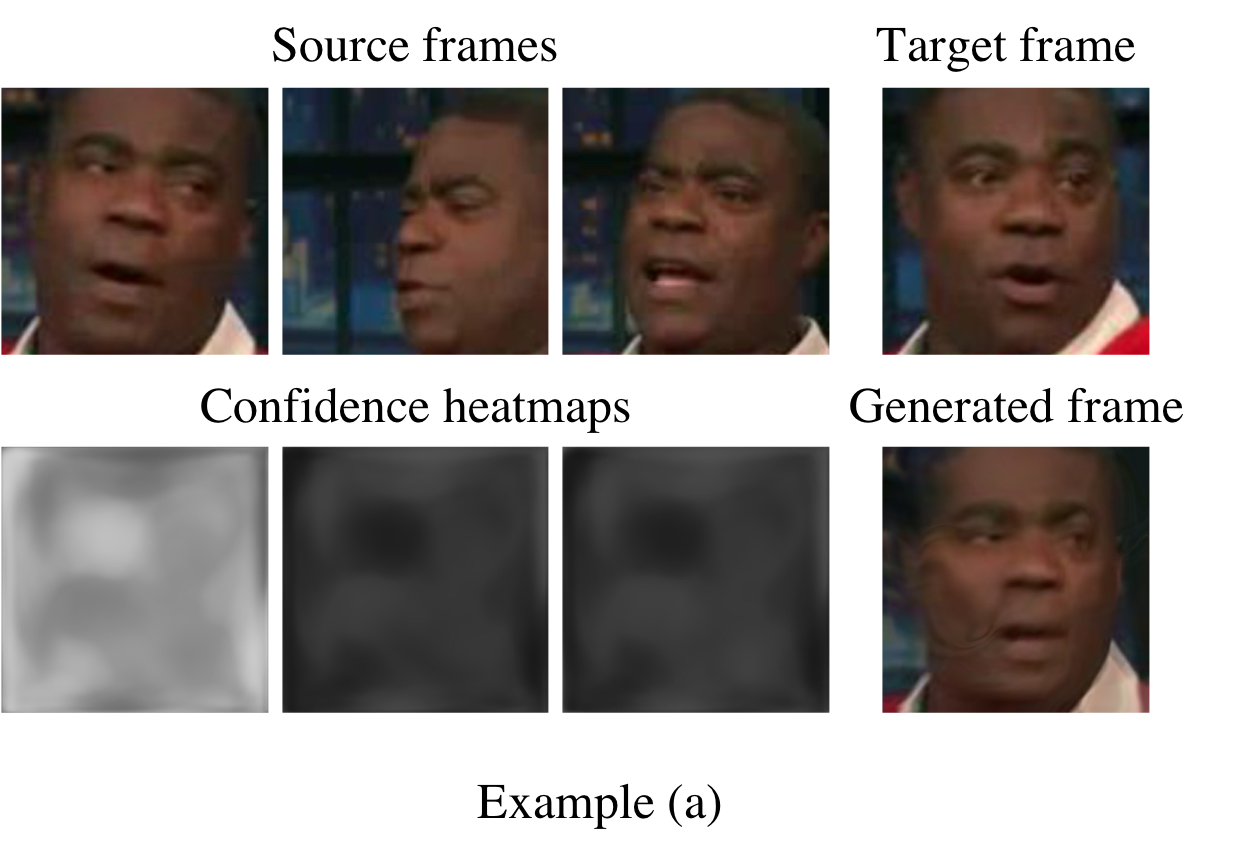} \\
\centering

\end{minipage}
\hspace{1cm}
\begin{minipage}[t][0.56\width]{0.4\textwidth}%
\centering

\includegraphics[width=\linewidth]{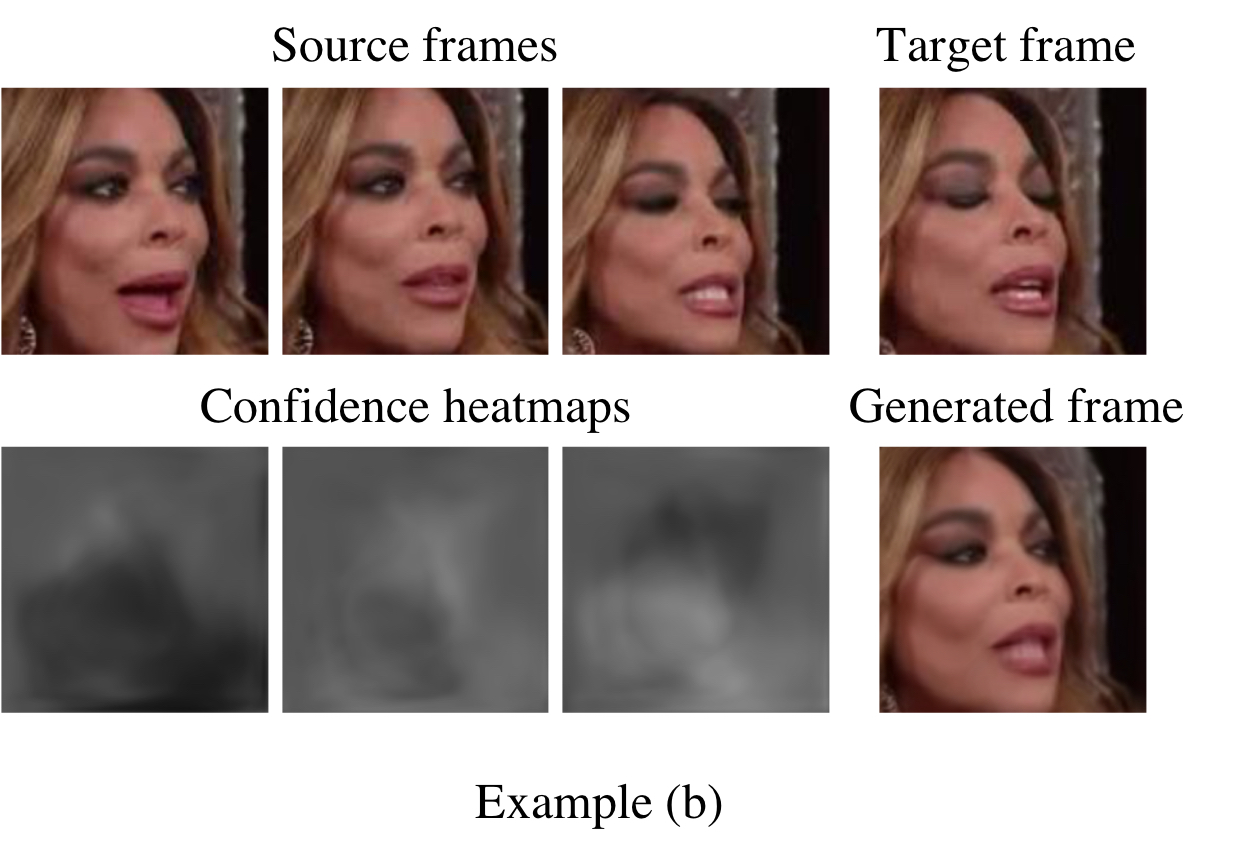} \\

\end{minipage}

    \vspace{-0.2cm}
    \caption{Confidence heatmaps learned by \networkname. Higher intensity corresponds to higher confidence. The network chooses the frames with most similar poses to draw from and ignores those with less similar poses  (see Example (a)). In Example (b), the mouth has higher confidence in the third source frame allowing the network to re-construct the teeth that are present in the target frame. More examples can be found in the supp. material.
    \vspace{-0.3cm}
    }
    \label{fig:attentionheatmaps}
\end{figure*}

\subsection{What is the benefit of additional source frames?}
\label{exp:additionalsourceframes}
The previous sections have shown that using additional source frames improves performance. 
This is at the expense of performing additional forward passes through the encoder (in this case two). 
Given enough GPU memory, these forward passes can be done in parallel, affecting only the memory requirements and not
the computational speed.

Using multiple source frames is further investigated by visualising confidence heatmaps for a given set of source frames  in \figref{fig:attentionheatmaps}.
The confidence heatmaps allow images with more similar pose to be used for creating the generated frame. Furthermore, the network can focus on one frame for generating a part of the face (e.g.~the mouth) and on another one for a different part.

\subsection{Image retrieval}
\label{exp:informationretrieval}
\begin{figure*}[t]
\begin{minipage}[t][1\width]{0.15\textwidth}%
\centering
\scriptsize
\raisebox{0.4cm}{(a)} {\includegraphics[height=0.8cm]{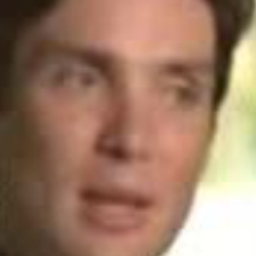}} \\ \vspace{.5em}
\raisebox{0.4cm}{(b)} {\includegraphics[height=0.8cm]{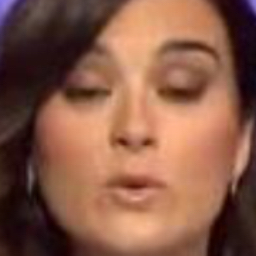}} \\ \vspace{1em}
\raisebox{0.4cm}{(c)} {\includegraphics[height=0.8cm]{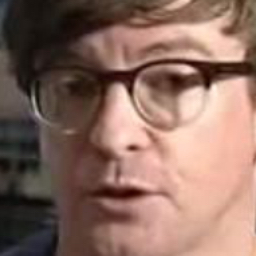}} \\ 
{      Query Images}
\end{minipage}
\begin{minipage}[t][1\width]{0.33\textwidth}%
\centering
\scriptsize
{\includegraphics[height=0.8cm,trim={9cm 0 0 0},clip]{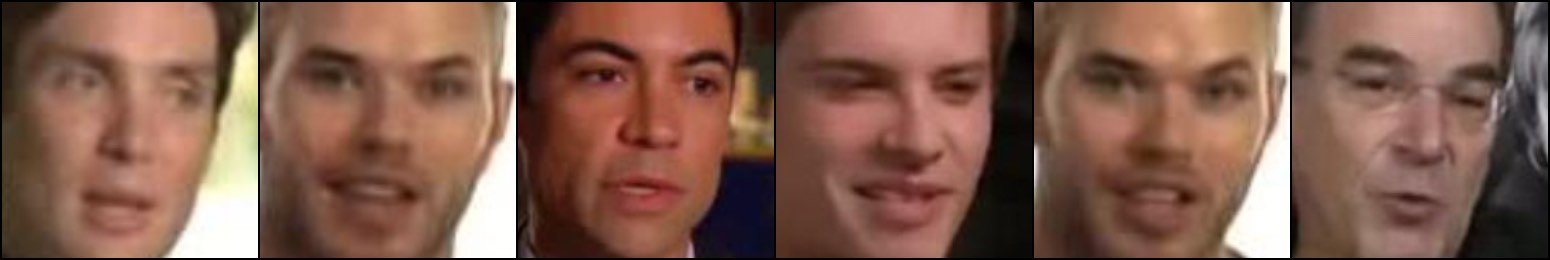}} \\ \vspace{.5em}
{\includegraphics[height=0.8cm,trim={9cm 0 0 0},clip]{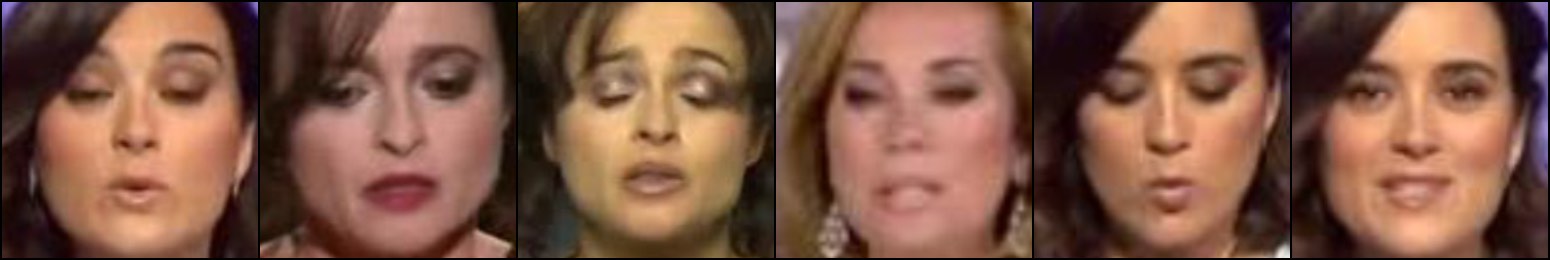}} \\ \vspace{1em}
{\includegraphics[height=0.8cm,trim={9cm 0 0 0},clip]{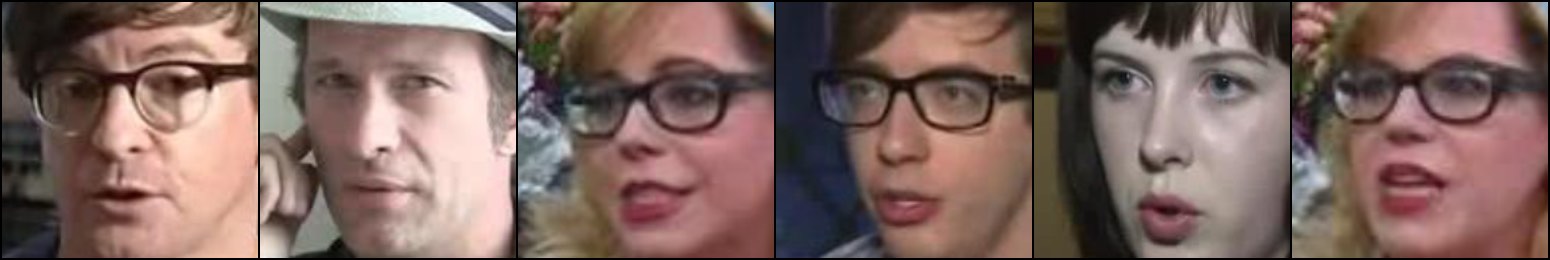}} \\
Nearest Neighbours. Arranged left to right.
\end{minipage}
\begin{minipage}[t][1\width]{0.15\textwidth}%
\centering
\scriptsize
\raisebox{0.4cm}{(d)} {\includegraphics[height=0.8cm]{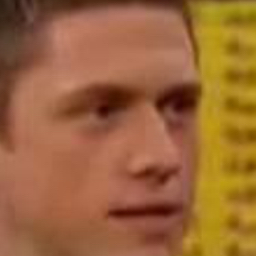}} \\ \vspace{.5em}
\raisebox{0.4cm}{(e)} {\includegraphics[height=0.8cm]{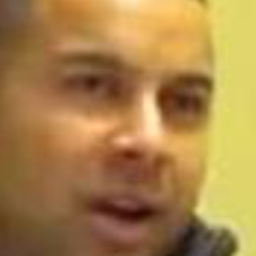}} \\ \vspace{1em}
\raisebox{0.4cm}{(f)} {\includegraphics[height=0.8cm]{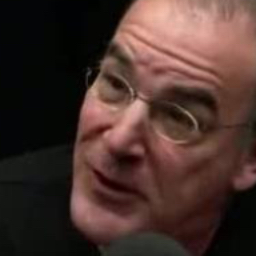}} \\ 
{      Query Images}
\end{minipage}
\begin{minipage}[t][1\width]{0.33\textwidth}%
\centering
\scriptsize
{\includegraphics[height=0.8cm,trim={9cm 0 0 0},clip]{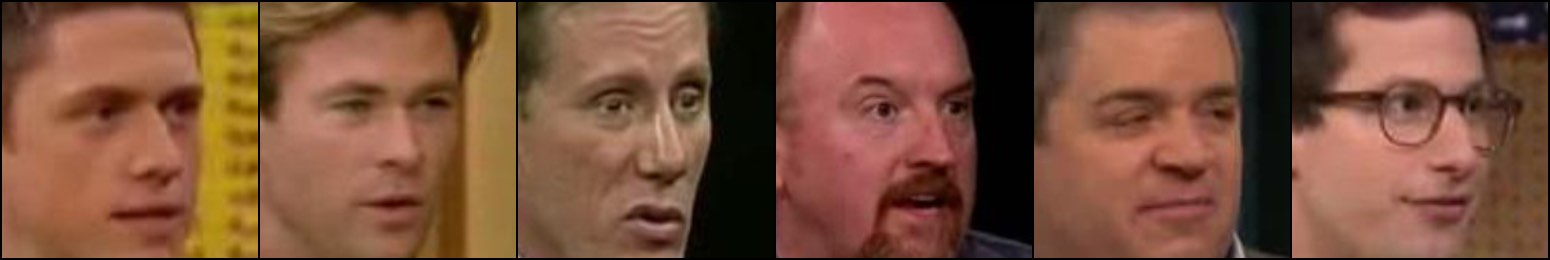}} \\ \vspace{.5em}
{\includegraphics[height=0.8cm,trim={9cm 0 0 0},clip]{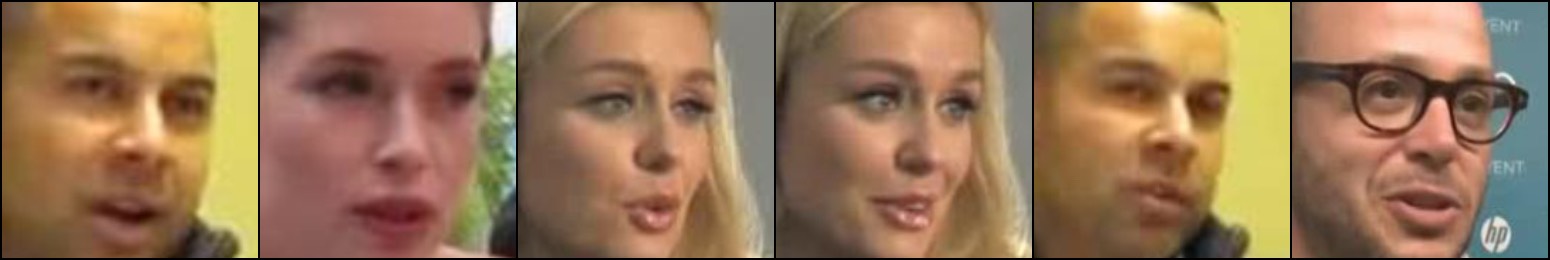}} \\ \vspace{1em}
{\includegraphics[height=0.8cm,trim={9cm 0 0 0},clip]{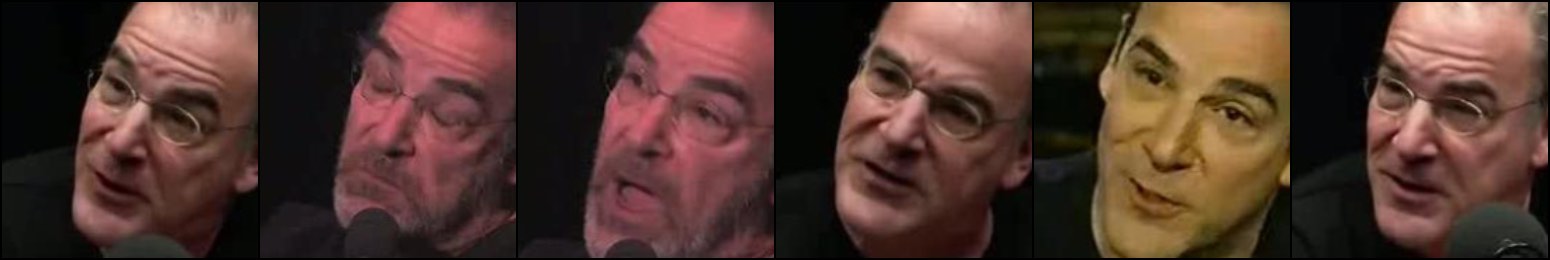}} \\
Nearest Neighbours. Arranged left to right.
\end{minipage}
\caption{Retrieval results using the embedding learned by \networkname. The embedding captures similar visual attributes since gallery images with similar facial attributes are retrieved for a given query image.
The retrieved images have similar pose to the query in all cases, and
the expression similarity can be seen for example in (b) 
with the eyes shut, and (a) with the  mouth slightly open. Please refer to the supp.~material for additional examples.}
\label{fig:retrievalexamples}
\end{figure*}

This section considers an application of the learned embedding:
retrieving images with similar facial attributes (e.g.~pose) but
across different identities.  To perform this task, a subset of 10,000
randomly sampled test images from \voxcombiname~is obtained.  For a
given query image, all other images (the gallery) are ranked based on
their similarity to the query image using the cosine similarity metric
between the corresponding embeddings.  For a given query image $Q$,
the embedding $x_q$ is extracted by performing a forward pass through
the network.  Similarly, the embedding $x_i$ is extracted for each
image $I_i$ in the gallery.  Each image $I_i$ is then ranked according
to the cosine similarity between $x_q$ and $x_i$.  If the network does
indeed encode salient information about facial attributes, the cosine
similarity can be used to identify images with similar poses and
facial attributes.  For a set of query images, the results are
visualised in \figref{fig:retrievalexamples}.  From these results it
is again affirmed that our embedding encodes information about facial
attributes, as the retrieved images have poses and expressions similar
to those of the query images.
Note, the embedding is largely unaffected by facial
decorations (e.g.\ glasses) and identity, as these do not change within a face-track and so do not need to be learned in order to predict the transformation.

\section{Conclusion}
We have introduced \networkname: a self-supervised framework for learning facial attributes from videos.
Our method learns about pose and expression by watching faces move and change over a large number of videos without {\em any} hand labels.
The features of our trained network can then be used to predict pose, landmarks, and expression on other datasets (despite the domain shift) by just training a linear layer on top of the learned embedding.
The features have been shown to be comparable or superior performance to self-supervised and supervised methods on a variety of tasks.
This is impressive as generally the performance of self-supervised methods has been found to be worse than that of supervised methods, yet our method is indeed competitive/superior to supervised methods for pose regression and facial landmark detection, and approaches supervised performance on expression classification.

\section{Acknowledgements}
The authors would like to thank James Thewlis for helpfully sharing code and datasets.
This work was funded by an EPSRC studentship and EPSRC Programme
Grant Seebibyte EP/M013774/1.

\bibliography{./shortstrings,./vgg_local,./vgg_other,./egbib}
\end{document}